\documentclass{article}
\usepackage[final]{colm2026_conference}
\usepackage{microtype}
\usepackage{hyperref}
\usepackage{url}
\usepackage{booktabs}
\usepackage{graphicx}
\usepackage{amsmath}
\usepackage{amssymb}
\usepackage{multirow}
\usepackage{xcolor}
\usepackage{subcaption}
\usepackage{pifont}
\usepackage{placeins}
\usepackage{adjustbox}
\usepackage{tikz}
\usepackage{enumitem}

\definecolor{darkblue}{rgb}{0, 0, 0.5}
\hypersetup{colorlinks=true, citecolor=darkblue, linkcolor=darkblue, urlcolor=darkblue}
 
\newcommand{\ci}{\textsc{ci}}
\newcommand{\sd}{\textsc{sd}}
\newcommand{\mm}{\textsc{mm}}
\newcommand{\ck}{\textsc{ck}}
\newcommand{\bci}{\textsc{bci}}
\newcommand{\cmark}{\ding{51}}
\newcommand{\xmark}{\ding{55}}
 
\title{The Story Shapes the Agent:
Narrative Priors in LLM Behavior}
 
\author{Yixuan Wang \\
University of North Carolina at Chapel Hill \\
\texttt{yxwang@cs.unc.edu}
\And
James Lester \\
North Carolina State University \\
\texttt{lester@ncsu.edu}
\And
Shashank Srivastava \\
University of North Carolina at Chapel Hill \\
\texttt{ssrivastava@cs.unc.edu}
}
 
\begin{document}
 
\ifcolmsubmission
\linenumbers
\fi
 
\maketitle

\begin{abstract}
Persona prompting is widely used to steer LLM agent behavior~\citep{tseng2024two, chen2024persona}, yet the narrative framing of a task can matter more than the assigned persona. We isolate this effect through structural isomorphism, constructing three text-based investigation games that share the same action space, stage progression, and resource constraints while varying only task narrative: disease investigation, IT troubleshooting, and murder mystery. Across 1,890 sessions spanning 3 models and 10 personas, we identify \textit{narrative priors}: systematic action tendencies activated by a task’s story framing, independent of its decision structure. Narrative priors explain 5--31$\times$ more behavioral variance than persona, are consistent across model architectures, and in two of three domains are negatively associated with task success. Persona effects 
that do transfer across narratives arise from \textit{behavioral anchors}, persona descriptions whose language maps directly onto shared actions. Causal interventions confirm this: removing anchor words from a high-transfer persona reduces cross-narrative consistency by 95\%. Our framework also generalizes to a held-out fourth narrative and yields a persona-selection method that improves cross-narrative transfer. These results suggest that LLM behavior that survives narrative changes should be grounded in concrete actions rather than abstract descriptions.
 
\end{abstract}

\section{Introduction}
\label{sec:intro}

A practitioner building an LLM-based agent faces a natural question: how do I make it behave the way I want? The dominant lightweight answer is \emph{persona prompting}: assign a dispositional description such as ``methodical and thorough'' or ``prefers learning through conversation,'' and expect the agent to behave accordingly. This expectation rests on an untested assumption: that persona-induced behavior will transfer when the same agent is redeployed in a different task domain. Prior work has studied whether LLMs can \textit{maintain} an assigned persona within a conversation~\citep{frisch2024llm, wang2024incharacter, samuel2024personagym}, but not whether the behavior generalizes to different task domains.

We test this assumption with a sharper question: when an agent’s behavior changes across task domains, is the persona adapting or is another process taking over? To answer this requires holding decision structure constant while varying only the surface-level story the agent is given. We implement this \emph{structural isomorphism} through three text-based investigation games that share identical action spaces, stage progressions, and resource constraints but differ in narrative: disease investigation, IT troubleshooting, and murder mystery (Figure~1). This design isolates narrative as a behavioral factor rather than confounding it with task structure. Running 10 learning-style personas across all three games on 3 models produces 1,890 sessions in which persona and narrative effects can be cleanly separated.

We find that task narrative is not merely a confound; it is the primary driver of agent behavior, explaining 5--31$\times$ more variance than the assigned persona. We term these implicit biases \textit{narrative priors}: behavioral tendencies that LLM agents inherit from pretraining corpora, activated by the story framing of a task independently of its decision structure. Medical narratives elicit more social interaction, IT narratives more document reading, crime narratives more diagnostic testing, mirroring domain stereotypes in pretraining text. Critically, these priors are not strategic adaptations: in two of three narratives, the narrative-biased action is negatively correlated with task success (\S\ref{sec:narrative_dom}).

Personas are still powerful. Within a fixed narrative, they explain substantial behavioral variance and can produce large differences in success. But their influence is narrative-confined: a persona that shapes behavior in one game often reshapes itself in another, and cross-narrative transfer classification averages 10.4\%, barely above the 10\% chance baseline. The persona effects that do survive narrative change are explained by what we call \emph{behavioral anchors}: descriptions whose concrete action language (e.g., ``prefers conversation'' $\rightarrow$ \texttt{talk}) binds directly to the shared action space, producing portable behavioral signatures. A causal intervention confirms this mechanism: abstracting away the anchor words while preserving the persona's psychological content reduces cross-narrative consistency by 95\% (\S\ref{sec:causal_anchors}).

These findings reframe how to design reliable LLM agents. Persona prompting works, but only within the behaviors the narrative already specifies. For behavioral control that survives task changes, personas must be grounded in concrete actions rather than abstract traits. More broadly, our structural isomorphism framework diagnoses narrative influence, our behavioral anchor analysis reveals which personas are robust to it, and our persona-selection method generalizes to a held-out fourth narrative domain, improving cross-narrative identifiability in all evaluation settings (\S\ref{sec:selection}, \S\ref{sec:cooking}).

\vspace{-1em}
\begin{figure}[t]
\centering
\begin{tikzpicture}
\node[inner sep=0pt, rounded corners=8pt, clip] {
  \includegraphics[width=0.75\textwidth]{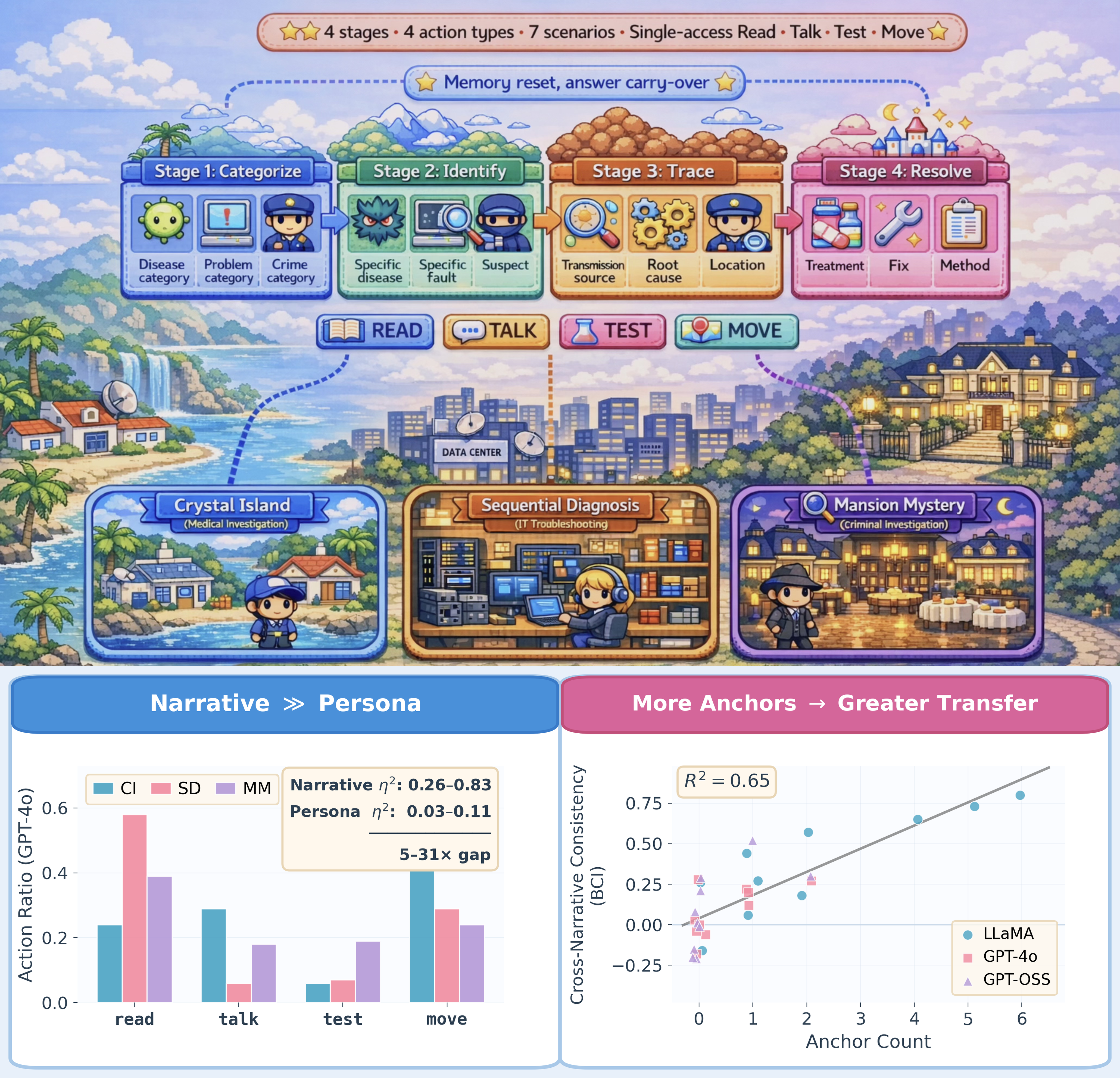}
};
\end{tikzpicture}
\caption{\textbf{Top:} Three structurally isomorphic interactive narrative games share identical decision structure (4 stages, 4 action types, 7 scenarios) but differ only in surface narrative (domain-level framing, e.g., medical vs.\ IT vs.\ crime). \textbf{Bottom left:} Despite shared structure, LLM agents exhibit very different action profiles across narratives, with narrative explaining 5--31$\times$ more variance than persona. \textbf{Bottom right:} Personas with more behavioral anchors (concrete action words) achieve higher cross-narrative consistency ($R^2 = 0.65$).}
\label{fig:overview}
\end{figure}

\section{Related Work}
\label{sec:related}

\noindent \textbf{Persona and Role Prompting.}
Persona prompting is a standard interface for shaping LLM behavior
 in multi-agent simulation \citep{park2023generative,
park2024generative}, dialogue personalization
\citep{zhang2018personalizing, shao2023character}, and role-based task
assignment \citep{xu2024character, wang2024rolellm}. Surveys
\citep{tseng2024two, chen2024persona} distinguish 
role-playing (assigning personas to LLMs) from personalization (adapting
LLMs to user profiles). A shared premise across this body of work is that assigned personas reliably govern behavior. We test that assumption by comparing persona influence to the task's narrative framing.

\noindent \textbf{Persona Consistency and Evaluation.}
Recent work has shown that LLM personas are more fragile than
often assumed: models fail to maintain personas under prompt
rephrasing \citep{gupta2023bias}, in multi-agent debate
\citep{baltaji2024conformity}, across extended dialogue
\citep{frisch2024llm}, and under adversarial probing
\citep{wang2024incharacter}. Recent benchmarks 
\citep{tu2024charactereval,samuel2024personagym}
measure persona consistency via linguistic consistency and
knowledge adherence. We instead ask whether persona-induced behavior \textit{transfers} across task contexts, and pinpoint the factors  responsible when it does not.

\noindent \textbf{Prompt Sensitivity and Framing Effects.}
LLMs are broadly sensitive to surface-level prompt variation: instruction
phrasing affects task performance \citep{lu2022fantastically,
webson2022prompt}, output format alters reasoning accuracy
\citep{zhao2021calibrate}, and example ordering changes few-shot results
\citep{lu2022fantastically}. Recent work on cognitive biases in LLMs
\citep{jones2022capturing} suggests that these sensitivities may reflect
systematic priors rather than random noise. We extend this line of
inquiry from textual sensitivity in static tasks to behavioral
sensitivity in interactive agent settings, showing that narrative framing
activates domain-specific behavioral priors that override 
persona instructions.

\noindent \textbf{LLM Agents in Interactive Environments.}
LLM-based agents have been deployed in text-based games
\citep{shridhar2020alfworld, trivedi2024appworld}, social simulations
\citep{park2023generative, park2024generative}, and multi-agent
workflows \citep{qian2024chatdev, hong2023metagpt}. Narrative-centered
learning environments have also served as testbeds for studying player
decision-making and engagement \citep{rowe2011integrating, wang2018high}.
While this literature focuses primarily on agent \textit{capabilities},
the question of how task framing shapes agent behavior independently
of task structure remains largely unexplored.

\section{Experimental Framework}
\label{sec:framework}

Most behavioral studies of LLM agents conflate \emph{what} the agent must decide with \emph{how} the task is described. A medical investigation and an IT troubleshooting scenario differ in action spaces, information structures, and success criteria, so behavioral differences are hard to interpret: they may reflect the decision problem, the narrative wrapper, or both. To identify \emph{narrative priors} rather than domain-specific behavior, we need tasks that vary in narrative surface but share the same decision-theoretic structure. We call this principle \textit{structural isomorphism} and implement it through three text-based investigation games.

\subsection{Isomorphic Game Design}
\label{sec:games}

All three games share a skeleton of structural invariants. Each consists of four investigation stages completed in fixed order, with per-stage turn limits. Between stages, the agent's context is cleared and only the answer from the preceding stage carries forward, preventing cumulative advantages from compounding across the investigation. Every information source (a document, a character, or a diagnostic test) may be accessed at most once per stage, forcing selectivity rather than exhaustive search. At each turn, the agent chooses from a discrete menu offering the same four core action types: \texttt{read} (consult documents), \texttt{talk} (interact with characters), \texttt{test} (run diagnostics), and \texttt{move} (navigate between locations). Options are pre-labeled by the game engine, so no manual or post-hoc action annotation is required. Across environments, the four stages map onto the same abstract decision sequence: $
\textit{categorize} \rightarrow \textit{identify} \rightarrow \textit{trace} \rightarrow \textit{resolve}$.
What differs is the surface narrative wrapper around the shared decision scaffold.


\textbf{Crystal Island (\ci{})} places the agent on a tropical research island in the role of a student investigating a disease outbreak. Across four stages the agent identifies the disease category, the specific disease, the transmission source, and the appropriate treatment. Our environment builds on the Crystal Island framework~\citep{rowe2011integrating}, originally designed for AI in education research. We extend the codebase with multi-stage investigation, stage-level memory reset, single-access constraints, and automated behavioral logging.

\textbf{Sequential Diagnosis (\sd{})} recasts the same structure in a corporate IT environment. The agent plays a support engineer who receives an urgent ticket and must identify the problem category, the specific fault, the root cause, and the fix.

\textbf{Mansion Mystery (\mm{})} recasts the same structure as a detective scenario. The agent plays an investigator solving a diamond theft in a Victorian mansion, determining the crime category, the suspect, the location, and the method.

All three games share 4 stages, 6 locations, 4 action types, and 7 scenarios. Minor differences in character count (5--7) and document count (12--15) reflect narrative needs rather than structural asymmetry; crucially, these resources are functionally equivalent, each providing single-access information within the same action-type category (environment property details in Appendix~\ref{app:env}). Thus, any behavioral differences across environments must stem not from different action inventories or investigation scaffolds, but from how the same scaffold is interpreted under different narrative frames.

\subsection{Persona Design}
\label{sec:personas}

We define 10 personas grounded in learning-style theory from educational psychology \citep{kolb2014experiential, felder1988learning}. The choice of \textit{personality-level descriptions} is deliberate. Role-based personas (e.g., ``you are a cautious analyst'') and explicit behavioral directives (e.g., ``always read before talking'') would make transfer trivial and reveal little about implicit narrative effects. Our design instead mirrors a realistic deployment scenario in which practitioners assign personality traits and expect those traits to shape behavior across task domains.

The 10 personas vary along four dimensions: processing speed (\textit{quick\_intuitive} vs.\ \textit{methodical\_thorough}), source preference (\textit{social\_collaborative} vs.\ \textit{coverage\_focused}), decision confidence (\textit{confident\_decisive} vs.\ \textit{verification\_seeking}), and investigation strategy (\textit{hands\_on\_practical}, \textit{hypothesis\_driven}, \textit{big\_picture\_conceptual}, \textit{effortful\_learner}). Table~\ref{tab:personas} lists each persona's core tendency alongside the action type most directly implied by its description, a distinction that becomes central to the analysis
of cross-narrative transferability (\S\ref{sec:anchors}).

Each persona prompt has three components: (1) a personality description ($\sim$80 words) identical across narratives; (2) a stage-specific behavioral reminder of 1--2 sentences; and (3) task instructions describing the agent's role, available resources and constraints. Only the third component changes across narratives, and describes available actions non-prescriptively. See Appendix~\ref{app:personas} for persona descriptions.

\begin{table}
\centering\footnotesize
\begin{tabular}{llc}
\toprule
\textbf{Persona} & \textbf{Core tendency} & \textbf{Action} \\
\midrule
quick\_intuitive      & Skims, trusts patterns       & -- \\
methodical\_thorough  & Reads carefully, verifies     & \texttt{read} \\
social\_collaborative & Conversation-first            & \texttt{talk} \\
coverage\_focused     & Covers every source once      & -- \\
confident\_decisive   & Commits quickly               & -- \\
verification\_seeking & Seeks convergent evidence     & -- \\
hands\_on\_practical  & Prefers doing and testing     & \texttt{test} \\
hypothesis\_driven    & Guesses then tests            & ${\sim}$\texttt{test} \\
big\_picture\_concept.& High-level patterns           & -- \\
effortful\_learner    & Reads slowly, persists        & \texttt{read} \\
\bottomrule
\end{tabular}
\caption{The 10 personas. \textit{Action} = action type most directly
implied by the description; ``--'' = no single action implied. The
relation to cross-narrative consistency is examined in
\S\ref{sec:anchors}.}
\label{tab:personas}
\end{table}

\subsection{Models and Configuration}
\label{sec:config}

We evaluate three models spanning different architectures and scales: LLaMA-3.1-70B (open-source), GPT-4o (proprietary), and GPT-4o-OSS-120B (open-source). This allows us to test whether narrative priors are architecture-specific or reflect shared properties of large-scale pretraining. For each model--narrative pair, we run all 10 personas, each across 7 scenarios with
3 independent trials, $10 \text{ personas} \times 7 \text{ scenarios} \times 3 \text{ trials} \times 9 \text{ model-environment pairs} = 1,890$ sessions. Each trial is an independent API session with no
shared state, and a sampling temperature of $T=0.7$.

\subsection{Behavioral Features}
\label{sec:features}

For each session we extract 16 normalized behavioral features organized in six categories: 

\begin{enumerate}[leftmargin=1em, topsep=0pt, itemsep=0pt]
    \item \textbf{Speed \& efficiency}: average turns per stage; exploration efficiency (unique locations\,/\,total moves).
    \item \textbf{Information gathering}: \texttt{read}, \texttt{talk}, and \texttt{test} ratios; information depth (information actions\,/\,unique sources).
    \item \textbf{Exploration pattern}: \texttt{move} ratio; location coverage; revisit rate.
    \item \textbf{Social behavior}: social breadth (unique characters interacted with\,/\,available); social dependency (\texttt{talk}\,/\,information actions).
    \item \textbf{Decision patterns}: action diversity; decision consistency; hesitation index.
    \item \textbf{Thoroughness}: resource coverage (unique resources\,/\,total available); experimentation rate (\texttt{test}\,/\,information actions).
\end{enumerate}

Four of the 16 features are computed as proportions of total core action types:
$\text{ratio}_a = \frac{n_a}{n_{\texttt{read}} + n_{\texttt{talk}} +
n_{\texttt{test}} + n_{\texttt{move}}}$. These ratios sum to 1, giving a compositional snapshot of each agent's action type profile. The remaining 12 features describe how an agent navigates and decides rather than which actions it takes (e.g., exploration efficiency, decision consistency, resource coverage). This ensures that the narrative dominance we report in \S\ref{sec:narrative_dom} is not driven solely by action-type encoding. 

These features serve two roles. First, they provide a measurement language for describing how behavior differs across personas and narratives. Second, they define a behavioral space in which we measure transfer. Full feature definitions are provided in Appendix~\ref{app:features}.

\section{Results}
\label{sec:results}

\subsection{Task Narrative Dominates Behavior}
\label{sec:narrative_dom}

The first question is straightforward: between persona and task
narrative, which one actually controls what the agent does?

\begin{figure}[t]
\centering
\includegraphics[width=\textwidth]{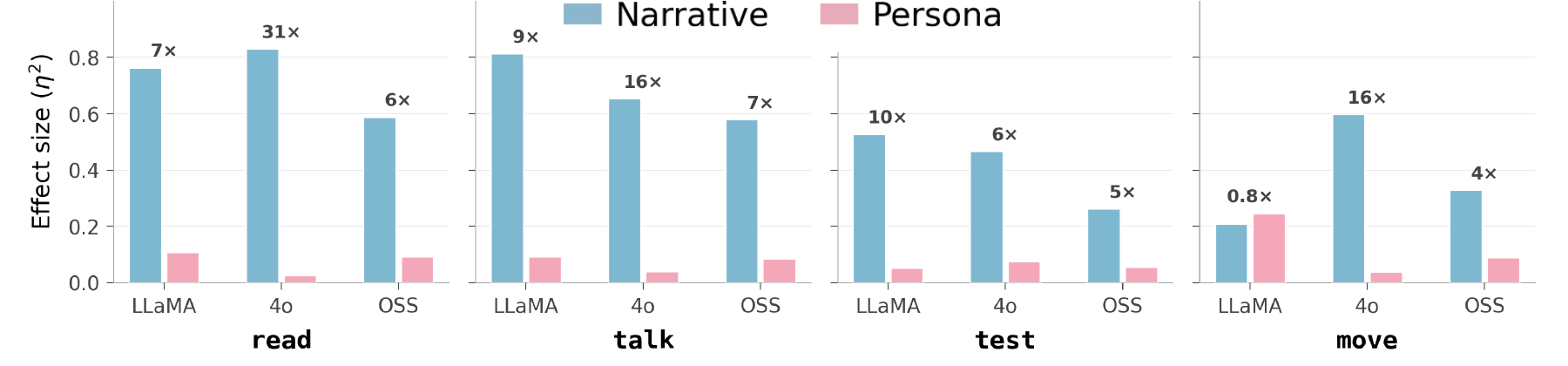}
\caption{Two-way ANOVA effect sizes ($\eta^2$) for Narrative and Persona
on the four core action ratios. Labels show the ratio of narrative to
persona $\eta^2$. Task narrative explains 5--31$\times$ more variance
than persona for information-gathering actions (\texttt{read},
\texttt{talk}, \texttt{test}). The sole exception is \texttt{move} for
LLaMA, where persona slightly exceeds narrative (0.8$\times$).
Classification accuracy (Random Forest, 16 features): narrative
99.7--100\%, persona 24.6--41.3\%.}
\label{fig:anova}
\end{figure}

\noindent \textbf{Variance decomposition.}
A two-way ANOVA (Narrative $\times$ Persona) on trial-averaged data
(70 observations per model--narrative cell; $N = 630$ per model) reveals
a stark asymmetry. For the three information-gathering actions
(\texttt{read}, \texttt{talk}, \texttt{test}), task narrative explains
5--31$\times$ more variance than persona, with all narrative effects
highly significant ($p < 10^{-15}$). The single exception is the
\texttt{move} ratio for LLaMA, where persona slightly exceeds narrative
($\eta^2 = .25$ vs.\ .21). Persona effects are significant in every
case ($p < .05$); personas do shape behavior, but narrative overwhelms them (Figure~\ref{fig:anova}).

Random Forest classifiers trained on the full 16-feature profiles
reinforce this picture: task narrative is predicted at 99.7--100\%
accuracy across all models, while persona classification reaches only
24.6--41.3\% (chance: 33\% and 10\% respectively). This gap holds even
when action-type ratios are excluded (\S\ref{sec:robustness}). We note that ANOVA $\eta^2$ is the primary evidence for narrative dominance; this comparison is conservative for narrative, which has only 3 levels versus persona's 10 and would therefore be expected to explain less variance.

\noindent \textbf{Narrative priors are not strategic adaptations.}
The behavioral signatures induced by narrative are consistent across all
three model architectures: \sd{} elicits the highest \texttt{read}
ratio, \ci{} the highest \texttt{talk} ratio, and \mm{} the highest
\texttt{test} ratio (Appendix~\ref{app:additional}). This pattern
mirrors domain stereotypes in pretraining text: medical contexts
co-occur with consultation language, IT contexts with documentation,
crime contexts with forensic investigation.

If these patterns reflected rational adaptation, the narrative-biased
action should predict task success. It generally does not. Reading more
in \sd{} does correlate with success ($r \approx +.34$, $p < .001$
across all models), but the \texttt{talk} bias in \ci{} is negatively
correlated with success in 2 of 3 models ($r = -.29$, $p < .001$ for
GPT-4o), and the \texttt{test} bias in \mm{} is negatively correlated
in 2 of 3 models ($r = -.25$, $p < .001$ for GPT-OSS). Full
correlations are in Appendix~\ref{app:additional}. The story framing
of a task causes agents to adopt domain-stereotypical behaviors that,
more often than not, hurt their performance.

\noindent \textbf{Persona effects are narrative-confined.}
Within a fixed narrative, persona explains substantial variance (mean $\eta^2$ up to 0.64 in \sd{}) and drives success rates from 0\% to 100\% across personas ($\chi^2 = 92.3$, $p < .001$ for \ci{} in LLaMA). Within-narrative persona classification reaches 31.9\% (3.2$\times$ chance), with anchor-rich personas most identifiable (\textit{social\_collaborative}: 90.5\% in \sd{}; Appendix~\ref{app:confusion}). But this signal collapses across narrative boundaries: cross-narrative transfer averages 10.4\%, barely above the 10\% baseline. Personas shape behavior powerfully within a narrative; the narrative determines which behavioral repertoire the persona operates within.

\subsection{Cross-Narrative Persona Consistency}
\label{sec:cross_game}

We define the Behavioral Consistency Index (\bci{}) to measure which personas maintain recognizable signatures across narratives. For persona $p$ in
narrative $g$, let $\vec{b}_p^{(g)} \in \mathbb{R}^{16}$ be the mean
behavioral feature vector. We z-normalize within each narrative to
remove narrative-level shifts:
\begin{equation}
\hat{b}_{p,f}^{(g)} = (b_{p,f}^{(g)} -
\mu_f^{(g)})/\sigma_f^{(g)}
\end{equation}
The \bci{} for persona $p$ is the mean pairwise Pearson correlation of
z-normalized profiles across all $\binom{3}{2} = 3$ narrative pairs:
\begin{equation}
\bci_p = \frac{1}{3} \sum_{(g_i,\, g_j)}
\text{corr}\!\left(\hat{\vec{b}}_p^{(g_i)},\;
\hat{\vec{b}}_p^{(g_j)}\right)
\end{equation}
A persona with high \bci{} deviates from the narrative mean in the same
direction and by similar magnitudes across all three games; a persona
with \bci{} near zero or negative reshapes its behavioral profile with
each new narrative.

\begin{figure}[t]
\centering
\includegraphics[width=1\textwidth]{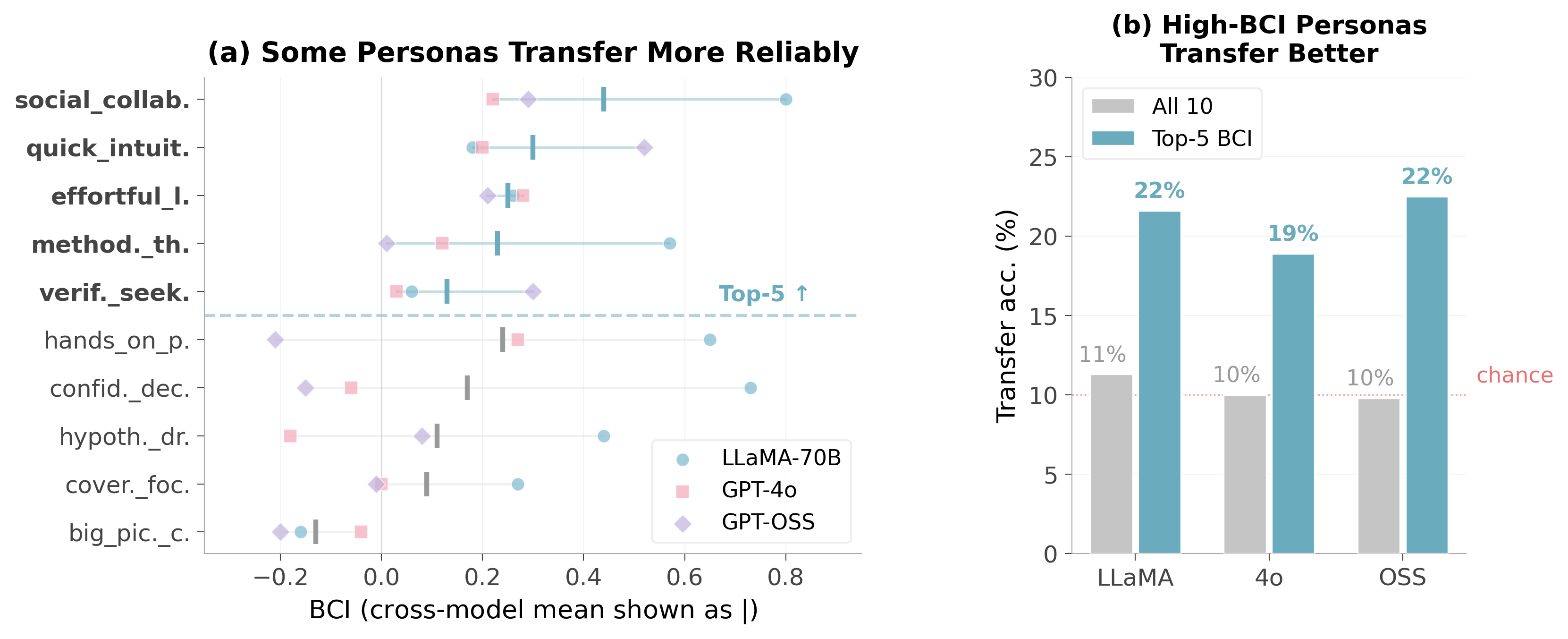}
\caption{\textbf{(a)} Behavioral Consistency Index (\bci{}) by persona.
Each point shows a single model; the vertical bar marks the cross-model
mean. The top-5 personas (above dashed line) maintain positive \bci{}
across all three models. \textbf{(b)} Cross-narrative classification
accuracy using all 10 personas vs.\ the top-5 \bci{} subset. High-\bci{}
filtering doubles transfer accuracy relative to the full set (chance =
10\%).}
\label{fig:bci}
\end{figure}

Figure~\ref{fig:bci}(a) reports the results. Five personas maintain positive
\bci{} across all three models, with \textit{social\_collaborative}
achieving the highest cross-model mean (+.44). At the other extreme,
\textit{big\_picture\_conceptual} is the only persona with consistently
negative \bci{} ($-.13$). A permutation null baseline (1,000 random
persona-label shuffles) places the 95th percentile at $+.33$, exceeded
by several persona--model combinations (marked with $\dagger$ in
Appendix~\ref{app:bci}).
Notably, \bci{} varies across models for the same persona (e.g.,
\textit{confident\_decisive}: $+.73$ in LLaMA, $-.15$ in GPT-OSS),
indicating that transferability is partially model-dependent.

We validate \bci{} through cross-narrative classification. Random Forest
classifiers trained on one narrative and tested on another achieve
10.4\% accuracy across all 10 personas. Restricting to the top-5 \bci{}
personas doubles this to 21.0\% (Figure~\ref{fig:bci}(b)), and the
advantage is fully preserved under a leave-one-narrative-out protocol
(+11.6pp; Appendix~\ref{app:additional}). \bci{} also shows negligible
correlation with human-rated prompt specificity ($\rho = 0.12$,
$p = 0.75$), ruling out prompt writing quality as a confound.

\subsection{What Explains \bci{}? Behavioral Anchors}
\label{sec:anchors}

What separates a persona that survives narrative change from one that dissolves into it? We hypothesize that the answer lies in \textit{behavioral anchors}: features that deviate consistently from the narrative-specific mean in the same direction across all narratives. Formally, feature $f$ is an anchor for persona $p$ if $|\hat{b}_{p,f}^{(g)}| > \tau$ with consistent sign across all $g$ ($\tau = 0.3$; robust across $[0.2, 0.5]$).

\textit{Social\_collaborative} in LLaMA has 6 anchors: ``prefers learning through conversation'' maps directly to \texttt{talk}, producing a stable signature across games. \textit{Big\_picture\_conceptual} has 0 anchors: ``high-level thinking'' lacks a stable action-level operationalization. Regression across all 30 model--persona pairs confirms this ($\bci_p = \alpha + \beta \cdot A_p + \epsilon$; $R^2 = 0.65$, $p < 10^{-7}$). But correlation does not establish causation; \S\ref{sec:causal_anchors} tests this directly.

\subsection{Persona Selection Algorithm}
\label{sec:selection}

The anchor analysis yields a practical method for selecting transferable personas without target-narrative data. Given behavioral data from two or more source narratives, we compute \bci{} from source data only, rank personas by source-\bci{}, and deploy the top-$K$ in the target narrative. Across all 9 model--target combinations with $K = 5$, source-\bci{} selection improves cross-narrative identifiability by +12.0pp on average, winning in all 9 conditions (Table~\ref{tab:selection} in Appendix~\ref{app:additional}). The improvement is monotonic in $K$: at $K = 3$, $+$25.4pp; at $K = 7$, $+$3.0pp. Source-only selection agrees with an oracle on 3.9/5 personas on average, with perfect agreement in 3 of 9 conditions.

\section{Intervention Experiments}
\label{sec:interventions}

Sections~\ref{sec:narrative_dom}--\ref{sec:selection} establish three empirical regularities: task narrative can shape behavior more strongly than persona, some personas transfer more reliably than others, and behavioral anchors strongly predict that transfer. But these analyses are primarily observational. We now ask three sharper questions. First, can a narrative prior be predicted in a previously unseen domain? Second, is the link between behavioral anchors and cross-narrative consistency genuinely causal? Third, does BCI capture something more informative than a simple text-level heuristic? We address each through experiments with GPT-4o-based agents.

\paragraph{Predicting Narrative Priors in a New Domain}
\label{sec:cooking}

If narrative priors reflect domain-level associations in pretraining corpora, then the behavioral bias of an unseen narrative should be predictable. We test this by constructing a fourth isomorphic game, \textbf{Cooking Kitchen (\ck{})}, which preserves all structural invariants. The agent plays a kitchen inspector investigating a dish preparation problem. Because cooking contexts often co-occurs with collaboration, coordination, and social activity, we pre-register the prediction that \ck{} will elicit a higher \texttt{talk} ratio than the mean of the original three narratives.

This prediction is confirmed: \texttt{talk} ratios form a graded hierarchy (\ci{} .286 $>$ \ck{} .213 $>$ \mm{} .183 $\gg$ \sd{} .062), with \ck{} significantly exceeding the 3-game mean of .177 ($t = 5.64$, $p < .001$, $d = 0.48$; Appendix~\ref{app:additional}). The \bci{} framework also generalizes cleanly: persona rankings are remarkably stable across 3 vs.\ 4 narratives ($\rho = 0.952$, $p < .001$), the top-5 persona set remains identical, and the anchor count regression improves slightly ($R^2 = 0.68$ vs.\ 0.65). These results suggest that both narrative priors and behavioral anchors capture regularities that extend beyond the initial three environments. 

\paragraph{Testing the Causal Role of Behavioral Anchors}
\label{sec:causal_anchors}

The anchor count--\bci{} regression ($R^2 = 0.65$) is observational. To test whether anchors \textit{cause} consistency, we directly manipulate anchor count while holding psychological content constant. We select the \textit{social\_collaborative} persona (\bci{} = +.44, 2 anchors: ``conversation'' and ``discussion'') and construct an abstract rewrite that preserves the collaborative disposition but removes concrete action words:

\vspace{0.3em}
\noindent\textit{Original:} ``Strongly prefers learning through
\underline{conversation} and \underline{discussion}. Naturally seeks
others' perspectives, starts by getting oriented through
\underline{conversation} before consulting other resources.''

\noindent\textit{Abstract:} ``Values collaborative epistemic exchange
and dialogic sense-making. Gravitates toward intersubjective knowledge
construction, preferring to establish shared understanding before
engaging with static information sources.''
\vspace{0.3em}

\noindent Both describe a collaborative learner who prioritizes shared
understanding before written sources; the difference is that the
original maps this disposition to a concrete action (\texttt{talk}),
while the abstract version does not. We run 63 sessions per version
(3 games $\times$ 7 scenarios $\times$ 3 trials). The effect is sharp. Removing the anchor words reduces \bci{} from +.22
to +.01, a 95\% drop (Table~\ref{tab:abstract_bci}). The collapse is
most dramatic in the \sd{}-\mm{} pair, where the original persona's
``conversation'' anchor created a stable social-action signal that
vanishes without the concrete word. At the feature level
(Table~\ref{tab:abstract_social}), the original persona maintains
consistently elevated \texttt{talk} behavior across all three
narratives ($z > +0.3$), while the abstract version's social signal is
substantially attenuated.

Multiple inferential checks support the same conclusion. A trial-level bootstrap (10,000 resamples) yields a $\Delta$\bci{} 95\% CI of $[-.38, -.04]$, excluding
zero. A Fisher $z$-transformation on the \sd{}-\mm{} pair yields
$z = 1.94$, $p = .026$ (one-tailed), and a permutation test (5,000
label shuffles) yields $p = .034$. The observed $\Delta$\bci{} of
$-0.21$ also matches the prediction from the anchor count regression
almost exactly: $\beta \times \Delta A = 0.11 \times (-2) = -0.22$. The
convergence of manipulation, bootstrapping, and regression prediction
provides strong evidence that behavioral anchors are causal.

We replicate this intervention on a second persona, \textit{coverage\_focused}, which uses concrete action vocabulary (``examine each available resource once'') but has a very different behavioral profile. Removing concrete action words via an abstract rewrite reduces \bci{} from $+.862$ to $+.370$ ($\Delta = -0.492$, permutation $p < 0.0001$, effect size $7.1$ SD vs.\ null), with the drop concentrating in CI-related pairs (CI-SD: $+.80 \to -.05$; CI-MM: $+.91 \to +.34$). This independently confirms that behavioral anchors are causally responsible for cross-narrative consistency.

\begin{table}[t]
\centering
\begin{minipage}[t]{0.48\textwidth}
\centering\small
\begin{tabular}{lccc}
\toprule
\textbf{Pair} & \textbf{Orig.} & \textbf{Abst.} & $\boldsymbol{\Delta}$ \\
\midrule
\ci{}-\sd{} & $-$.23 & $-$.18 & +.05 \\
\ci{}-\mm{} & +.13 & $-$.02 & $-$.15 \\
\sd{}-\mm{} & +.76 & +.22 & $-$.54 \\
\midrule
\textbf{\bci{}} & \textbf{+.22} & \textbf{+.01} & $\boldsymbol{-}$\textbf{.21} \\
\bottomrule
\end{tabular}
\captionof{table}{Pairwise correlations for
\textit{social\_collaborative}: original vs.\ abstract rewrite.
Removing anchors reduces \bci{} by 95\%.}
\label{tab:abstract_bci}
\end{minipage}\hfill
\begin{minipage}[t]{0.48\textwidth}
\centering\small
\begin{tabular}{lcc}
\toprule
\textbf{Narr.} & \textbf{Orig.\ $z_{\texttt{talk}}$} & \textbf{Abst.\ $z_{\texttt{talk}}$} \\
\midrule
\ci{} & +0.31 & +0.12 \\
\sd{} & +0.88 & +0.34 \\
\mm{} & +0.72 & +0.15 \\
\bottomrule
\end{tabular}
\captionof{table}{Z-normalized \texttt{talk} scores. The abstract
version's social signal largely disappears.}
\label{tab:abstract_social}
\end{minipage}
\end{table}

\paragraph{Behavioral vs.\ Textual Predictors of Transfer}
\label{sec:baseline}

A natural objection is that \bci{} may simply discover a superficial text-level heuristic: personas transfer when they contain more action-related words. We test this with an action-word (AW) baseline that counts manually specified action-linked words in each persona description (Appendix~\ref{app:aw_list}). \bci{} significantly predicts cross-narrative transfer accuracy ($\rho = 0.68$--$0.76$, $p < .05$), while AW does not ($\rho = 0.38$--$0.55$, $p > .10$). \bci{}-based top-5 outperforms AW-based top-5 by +4.8pp on average. Two concrete cases make this constrast intuitive: \textit{quick\_intuitive} contains zero action words yet ranks 2nd in \bci{} (its ``skim and move on'' style is behaviorally distinct), while \textit{coverage\_focused} contains 4 action words but ranks 9th (its words map to every action type, producing a flat profile). More fundamentally, \bci{} is model-adaptive (cross-model $\rho < 0.24$); a fixed text metric cannot capture these model-specific behaviors.

\paragraph{Robustness to Stronger Prompts and Abstract Labels}
\label{sec:directive_abstract}

Two additional experiments test the boundaries of narrative priors. First, we replace personality-level personas with explicit behavioral directives mandating at least 60\% of actions be a target type (\texttt{read\_directive}, \texttt{talk\_directive}, \texttt{test\_directive}; 63 sessions on GPT-4o). The directives are effective: target actions increase in 8 of 9 cells (Cohen $d$ up to $+5.05$). Yet the cross-narrative range of each directive's target ratio (0.17--0.27 across games) matches or exceeds the original cross-narrative range without any directive (0.13--0.33). Even under explicit behavioral mandates, narrative framing governs how much the agent complies. Notably, \texttt{test\_directive} on \mm{} fails entirely to raise the test ratio above its narrative baseline ($\Delta = -0.006$, $p = 0.71$), because \mm{}'s forensic-testing prior already saturates that action.

Second, we replace all action verbs with abstract codes (\texttt{read} $\to$ \texttt{ACTION\_ALPHA}, \texttt{talk} $\to$ \texttt{ACTION\_BETA}, etc.) via an API-layer wrapper (21 sessions on GPT-4o). None of the three narrative priors weaken; two intensify. In \ci{}, the \texttt{talk} ratio rises from .310 to .518 ($+67\%$, $p = .014$) while \texttt{read} collapses from .291 to .036. In \sd{}, the dominant \texttt{read} ratio is preserved at .587 ($p = .68$). In \mm{}, the \texttt{test} ratio rises from .178 to .319 ($+79\%$, $p = .27$ with high variance at $N = 7$). Removing verb-level cues makes the model rely more heavily on narrative context, ruling out the surface-semantics interpretation and localizing narrative priors to the domain framing itself.

\section{Robustness Analyses}
\label{sec:robustness}

We address five potential concerns. First, the four action-type ratios might trivially encode narrative identity; however, replicating all analyses with only the 12 non-ratio features yields unchanged narrative prediction accuracy (99.7--100.0\%) and strongly correlated \bci{} values (Spearman $\rho = 0.84$--$0.99$). Second, personas do affect more than action selection: 4--6 of 6 structural features show significant persona effects ($p < .05$, $\eta^2$ up to 0.29) within each narrative, including exploration patterns, decision consistency, and resource coverage. Third, personas are not simply ineffective prompts: within-narrative $\eta^2$ reaches 0.64, and per-game success rates range from 0\% to 100\% across personas. Fourth, anchor count is not confounded by prompt quality: it shows negligible correlation with prompt length ($\rho = 0.08$) and human-rated specificity ($\rho = 0.15$). Fifth, to verify that ANOVA results are not driven by distributional assumptions on bounded proportion data, we reran all 12 model $\times$ action-ratio comparisons under Kruskal-Wallis tests and binomial GLM with logit link. All narrative-dominance conclusions hold ($p < 10^{-9}$ in 11 of 12 cells), with effect-size ratios comparable to or larger than those under ANOVA.

\section{Discussion}
\label{sec:discussion}

\noindent \textbf{Narrative priors as implicit behavioral bias.}
Our central finding is that LLM agents exhibit \textit{narrative priors}: implicit behavioral biases activated by story framing, independent of decision structure. The biases are consistent across architectures ($r = .96$), hurt task success in two of three narratives, modulate structural features beyond action frequencies, and are predictable in a novel domain (\S\ref{sec:cooking}, $\rho = 0.952$). This poses a concrete reliability challenge. When a practitioner deploys the same persona across domains, the behavioral profile will be driven more by narrative than by the assigned persona, and the narrative-induced behaviors may actively hurt performance ($r = -.29$, $p < .001$ for \texttt{talk} in \ci{}).

\noindent \textbf{Behavioral anchors.}
Persona descriptions that create direct semantic--action bindings (``prefers conversation'' $\to$ \texttt{talk}) produce portable behavioral signatures; abstract descriptions do not. The causal experiment (\S\ref{sec:causal_anchors}) confirms this on two independent personas: removing concrete action words reduces \bci{} by 95\% for \textit{social\_collaborative} and by 57\% for \textit{coverage\_focused}, with the observed drops matching regression predictions. This yields a design principle: for cross-narrative consistency, personas should reference concrete actions rather than abstract dispositions. More broadly, it suggests that the effectiveness of natural-language behavioral instructions may depend critically on how directly they map onto the available action space.

\noindent \textbf{Practical implications.}
These findings yield concrete guidance for agent builders: to achieve portable behavior across task domains, persona descriptions should reference concrete actions (e.g., ``prioritize reading documents'') rather than abstract dispositions (e.g., ``be thorough''). Even explicit behavioral directives mandating specific action frequencies do not override narrative priors (\S\ref{sec:directive_abstract}), and narrative priors persist when action labels are fully abstracted, localizing their source to domain framing rather than action-verb semantics.

\noindent \textbf{Limitations.} Our narratives belong to the sequential investigation genre; extending the isomorphism framework to structurally different task families such as planning, tool-use, or multi-agent collaboration is an important next step. The 10 personas vary along learning-style dimensions; our supplementary directive experiment suggests narrative priors persist even under explicit behavioral mandates. We analyze action-level behavior rather than linguistic output; analyzing chain-of-thought content could reveal how narrative priors form at the generation level. Finally, \bci{} rankings show limited cross-model consistency ($\rho < 0.24$), suggesting persona selection may need per-model calibration.

Even with these limitations, the main message is clear: persona prompting is not a context-invariant control interface. The story wrapped around a task can shape agent behavior more strongly than the persona itself. Understanding, predicting, and mitigating that effect is likely to be important for building LLM agents with reliable behavior across domains. Our structural isomorphism framework provides a diagnostic for measuring this effect, and our behavioral anchor analysis offers a practical path toward personas that are robust to it.




\bibliographystyle{colm2026_conference}
\bibliography{references}

@inproceedings{park2023generative,
  title={Generative agents: Interactive simulacra of human behavior},
  author={Park, Joon Sung and O'Brien, Joseph and Cai, Carrie Jun and Morris, Meredith Ringel and Liang, Percy and Bernstein, Michael S},
  booktitle={Proceedings of the 36th annual acm symposium on user interface software and technology},
  pages={1--22},
  year={2023}
}

@article{park2024generative,
  title={Generative agent simulations of 1,000 people},
  author={Park, Joon Sung and Zou, Carolyn Q and Shaw, Aaron and Hill, Benjamin Mako and Cai, Carrie and Morris, Meredith Ringel and Willer, Robb and Liang, Percy and Bernstein, Michael S},
  journal={arXiv preprint arXiv:2411.10109},
  year={2024}
}

@inproceedings{zhang2018personalizing,
  title={Personalizing dialogue agents: I have a dog, do you have pets too?},
  author={Zhang, Saizheng and Dinan, Emily and Urbanek, Jack and Szlam, Arthur and Kiela, Douwe and Weston, Jason},
  booktitle={Proceedings of the 56th Annual Meeting of the Association for Computational Linguistics (Volume 1: Long Papers)},
  pages={2204--2213},
  year={2018}
}

@article{xu2024character,
  title={Character is destiny: Can large language models simulate personadriven decisions in role-playing},
  author={Xu, Rui and Wang, Xintao and Chen, Jiangjie and Yuan, Siyu and Yuan, Xinfeng and Liang, Jiaqing and Chen, Zulong and Dong, Xiaoqing and Xiao, Yanghua},
  journal={arXiv preprint arXiv:2404.12138},
  year={2024}
}

@inproceedings{wang2024rolellm,
  title={Rolellm: Benchmarking, eliciting, and enhancing role-playing abilities of large language models},
  author={Wang, Noah and Peng, Zy and Que, Haoran and Liu, Jiaheng and Zhou, Wangchunshu and Wu, Yuhan and Guo, Hongcheng and Gan, Ruitong and Ni, Zehao and Yang, Jian and others},
  booktitle={Findings of the Association for Computational Linguistics: ACL 2024},
  pages={14743--14777},
  year={2024}
}

@article{gupta2023bias,
  title={Bias runs deep: Implicit reasoning biases in persona-assigned llms},
  author={Gupta, Shashank and Shrivastava, Vaishnavi and Deshpande, Ameet and Kalyan, Ashwin and Clark, Peter and Sabharwal, Ashish and Khot, Tushar},
  journal={arXiv preprint arXiv:2311.04892},
  year={2023}
}

@inproceedings{tseng2024two,
  title={Two tales of persona in llms: A survey of role-playing and personalization},
  author={Tseng, Yu-Min and Huang, Yu-Chao and Hsiao, Teng-Yun and Chen, Wei-Lin and Huang, Chao-Wei and Meng, Yu and Chen, Yun-Nung},
  booktitle={Findings of the Association for Computational Linguistics: EMNLP 2024},
  pages={16612--16631},
  year={2024}
}

@article{chen2024persona,
  title={From persona to personalization: A survey on role-playing language agents},
  author={Chen, Jiangjie and Wang, Xintao and Xu, Rui and Yuan, Siyu and Zhang, Yikai and Shi, Wei and Xie, Jian and Li, Shuang and Yang, Ruihan and Zhu, Tinghui and others},
  journal={arXiv preprint arXiv:2404.18231},
  year={2024}
}

@inproceedings{baltaji2024conformity,
  title={Conformity, confabulation, and impersonation: Persona inconstancy in multi-agent LLM collaboration},
  author={Baltaji, Razan and Hemmatian, Babak and Varshney, Lav},
  booktitle={Proceedings of the 2nd Workshop on Cross-Cultural Considerations in NLP},
  pages={17--31},
  year={2024}
}

@inproceedings{frisch2024llm,
  title={LLM agents in interaction: Measuring personality consistency and linguistic alignment in interacting populations of large language models},
  author={Frisch, Ivar and Giulianelli, Mario},
  booktitle={Proceedings of the 1st Workshop on Personalization of Generative AI Systems (PERSONALIZE 2024)},
  pages={102--111},
  year={2024}
}

@book{kolb2014experiential,
  title={Experiential learning: Experience as the source of learning and development},
  author={Kolb, David A},
  year={2014},
  publisher={FT press}
}

@article{felder1988learning,
  title={Learning and teaching styles in engineering education},
  author={Felder, Richard M and Silverman, Linda K and others},
  journal={Engineering education},
  volume={78},
  number={7},
  pages={674--681},
  year={1988},
  publisher={North Carolina}
}

@inproceedings{tu2024charactereval,
  title={Charactereval: A chinese benchmark for role-playing conversational agent evaluation},
  author={Tu, Quan and Fan, Shilong and Tian, Zihang and Shen, Tianhao and Shang, Shuo and Gao, Xin and Yan, Rui},
  booktitle={Proceedings of the 62nd Annual Meeting of the Association for Computational Linguistics (Volume 1: Long Papers)},
  pages={11836--11850},
  year={2024}
}

@article{samuel2024personagym,
  title={Personagym: Evaluating persona agents and llms},
  author={Samuel, Vinay and Zou, Henry Peng and Zhou, Yue and Chaudhari, Shreyas and Kalyan, Ashwin and Rajpurohit, Tanmay and Deshpande, Ameet and Narasimhan, Karthik and Murahari, Vishvak},
  journal={arXiv preprint arXiv:2407.18416},
  volume={8},
  number={9},
  year={2024}
}

@inproceedings{wang2024incharacter,
  title={Incharacter: Evaluating personality fidelity in role-playing agents through psychological interviews},
  author={Wang, Xintao and Xiao, Yunze and Huang, Jen-tse and Yuan, Siyu and Xu, Rui and Guo, Haoran and Tu, Quan and Fei, Yaying and Leng, Ziang and Wang, Wei and others},
  booktitle={Proceedings of the 62nd annual meeting of the association for computational linguistics (volume 1: Long papers)},
  pages={1840--1873},
  year={2024}
}

@inproceedings{shao2023character,
  title={Character-llm: A trainable agent for role-playing},
  author={Shao, Yunfan and Li, Linyang and Dai, Junqi and Qiu, Xipeng},
  booktitle={Proceedings of the 2023 Conference on Empirical Methods in Natural Language Processing},
  pages={13153--13187},
  year={2023}
}

@inproceedings{lu2022fantastically,
  title={Fantastically ordered prompts and where to find them: Overcoming few-shot prompt order sensitivity},
  author={Lu, Yao and Bartolo, Max and Moore, Alastair and Riedel, Sebastian and Stenetorp, Pontus},
  booktitle={Proceedings of the 60th Annual Meeting of the Association for Computational Linguistics (Volume 1: Long Papers)},
  pages={8086--8098},
  year={2022}
}

@inproceedings{webson2022prompt,
  title={Do prompt-based models really understand the meaning of their prompts?},
  author={Webson, Albert and Pavlick, Ellie},
  booktitle={Proceedings of the 2022 conference of the north american chapter of the association for computational linguistics: Human language technologies},
  pages={2300--2344},
  year={2022}
}

@inproceedings{zhao2021calibrate,
  title={Calibrate before use: Improving few-shot performance of language models},
  author={Zhao, Zihao and Wallace, Eric and Feng, Shi and Klein, Dan and Singh, Sameer},
  booktitle={International conference on machine learning},
  pages={12697--12706},
  year={2021},
  organization={Pmlr}
}

@article{jones2022capturing,
  title={Capturing failures of large language models via human cognitive biases},
  author={Jones, Erik and Steinhardt, Jacob},
  journal={Advances in Neural Information Processing Systems},
  volume={35},
  pages={11785--11799},
  year={2022}
}

@article{shridhar2020alfworld,
  title={Alfworld: Aligning text and embodied environments for interactive learning},
  author={Shridhar, Mohit and Yuan, Xingdi and C{\^o}t{\'e}, Marc-Alexandre and Bisk, Yonatan and Trischler, Adam and Hausknecht, Matthew},
  journal={arXiv preprint arXiv:2010.03768},
  year={2020}
}

@inproceedings{trivedi2024appworld,
  title={Appworld: A controllable world of apps and people for benchmarking interactive coding agents},
  author={Trivedi, Harsh and Khot, Tushar and Hartmann, Mareike and Manku, Ruskin and Dong, Vinty and Li, Edward and Gupta, Shashank and Sabharwal, Ashish and Balasubramanian, Niranjan},
  booktitle={Proceedings of the 62nd Annual Meeting of the Association for Computational Linguistics (Volume 1: Long Papers)},
  pages={16022--16076},
  year={2024}
}

@inproceedings{qian2024chatdev,
  title={Chatdev: Communicative agents for software development},
  author={Qian, Chen and Liu, Wei and Liu, Hongzhang and Chen, Nuo and Dang, Yufan and Li, Jiahao and Yang, Cheng and Chen, Weize and Su, Yusheng and Cong, Xin and others},
  booktitle={Proceedings of the 62nd annual meeting of the association for computational linguistics (volume 1: Long papers)},
  pages={15174--15186},
  year={2024}
}

@inproceedings{hong2023metagpt,
  title={MetaGPT: Meta programming for a multi-agent collaborative framework},
  author={Hong, Sirui and Zhuge, Mingchen and Chen, Jonathan and Zheng, Xiawu and Cheng, Yuheng and Wang, Jinlin and Zhang, Ceyao and Wang, Zili and Yau, Steven Ka Shing and Lin, Zijuan and others},
  booktitle={The twelfth international conference on learning representations},
  year={2023}
}

@article{rowe2011integrating,
  title={Integrating learning, problem solving, and engagement in narrative-centered learning environments},
  author={Rowe, Jonathan P and Shores, Lucy R and Mott, Bradford W and Lester, James C},
  journal={International Journal of Artificial Intelligence in Education},
  volume={21},
  number={1-2},
  pages={115--133},
  year={2011},
  publisher={SAGE Publications Sage UK: London, England}
}

@inproceedings{wang2018high,
  title={High-Fidelity Simulated Players for Interactive Narrative Planning.},
  author={Wang, Pengcheng and Rowe, Jonathan P and Min, Wookhee and Mott, Bradford W and Lester, James C},
  booktitle={IJCAI},
  volume={18},
  number={1},
  pages={13--19},
  year={2018}
}
\newpage
\appendix

\section{Structural properties}
\label{app:env}

\begin{table}[h!]
\centering
\small
\label{tab:env_properties}
\begin{tabular}{lccc}
\toprule
Property & CI & SD & MM \\
\midrule
Stages & 4 & 4 & 4 \\
Locations & 6 & 6 & 6 \\
Characters & 7 & 5 & 7 \\
Documents & 12 & 12 & 15 \\
Core action types & 4 & 4 & 4 \\
Scenarios & 7 & 7 & 7 \\
\bottomrule
\end{tabular}
\caption{Structural properties of the three narrative environments. Core decision architecture is identical across environments; minor differences in resource counts reflect narrative packaging rather than changes to the underlying investigation structure.}
\end{table}


\section{Persona Descriptions}
\label{app:personas}

Each persona prompt consists of three components: (1) a personality
description ($\sim$80 words), (2) a stage-specific behavioral reminder,
and (3) task instructions (game rules, single-access constraints). Below
we provide the personality descriptions and reminders. The task
instructions are identical across personas and enforce the structural
invariants described in \S\ref{sec:games}. Across task narratives, only
the role label (student/engineer/detective) and domain-specific resource
names change; the personality content is held constant.

\paragraph{Quick Intuitive.} Learns quickly through pattern recognition.
Trusts first impressions, skims for key points, comfortable with partial
information. Prefers to gather information efficiently and move on rather
than lingering. \textit{Reminder: Trust your pattern recognition and move
efficiently. Skim for key patterns and connections.}

\paragraph{Methodical Thorough.} Careful, systematic, prefers solid
evidence before deciding. Reads important information carefully and
verifies key facts by consulting different source types. Balances
thoroughness with efficiency. \textit{Reminder: Be thorough on first pass,
focus careful attention where it matters most. Verify by consulting
different sources, not re-reading.}

\paragraph{Social Collaborative.} Strongly prefers learning through
conversation and discussion. Naturally seeks others' perspectives, starts
by getting oriented through conversation before consulting other resources.
Understands technical details eventually require written sources.
\textit{Reminder: Start with conversations to get oriented, then use other
resources for specific details.}

\paragraph{Effortful Learner.} Finds dense content genuinely challenging.
Reads slowly and carefully, focuses on main ideas and key facts. Persistent
in completing tasks even when content is difficult. Accepts partial
understanding rather than getting stuck. \textit{Reminder: Read carefully
on first pass, focus on main points, keep moving forward even with partial
clarity.}

\paragraph{Confident Decisive.} Trusts own judgment, commits quickly once
sufficient information is gathered. Does not second-guess or proactively
seek contradicting information. Willing to change mind if new information
contradicts conclusion, but does not dwell on uncertainty.
\textit{Reminder: Make clear decisions with confidence. Once decided, trust
your judgment and move forward.}

\paragraph{Verification-Seeking.} Wants convergent evidence from multiple
different sources before deciding. Seeks corroboration across diverse
source types. Cross-checks across sources, not within the same source.
\textit{Reminder: Look for convergence across different source types.
Cross-check across sources, not within the same source.}

\paragraph{Hands-On Practical.} Learns best by doing and testing. Prefers
direct verification over reading. Finds practical experimentation more
engaging than lengthy explanations. Reads when necessary but keeps it
brief. \textit{Reminder: Prefer hands-on investigation when possible.
Keep reading brief and move to practical action.}

\paragraph{Coverage-Focused.} Systematic and comprehensive in information
gathering. Wants to examine each available resource once but thoroughly.
Feels more confident when coverage is complete. Thoroughness means covering
many sources once, not re-examining. \textit{Reminder: Be systematic and
comprehensive. Cover all available resources once but thoroughly.}

\paragraph{Big-Picture Conceptual.} Focuses on overall concepts and
patterns rather than specific details. Skims fine details, good at seeing
connections between concepts. Sometimes misses important details but
understands overall logic. \textit{Reminder: Focus on overall concepts and
main ideas. Look for the big picture rather than getting caught up in
details.}

\paragraph{Hypothesis-Driven.} Forms quick hypotheses and tests through
action rather than extensive research. Comfortable with initial guesses,
sees incorrect attempts as informative. Adaptable and willing to change
mind based on feedback. \textit{Reminder: Form quick hypotheses and test
them. Gather enough to make a reasonable guess, then act.}

\paragraph{Domain Mapping.} Across task narratives, personas use the
following role and resource mappings: \ci{} uses a 10th-grade student
investigating disease (read books/posters, talk to NPCs, test items);
\sd{} uses a tech support engineer troubleshooting IT issues (read
logs/docs, talk to users/experts, run diagnostics); \mm{} uses a detective
solving a criminal case (read documents/evidence, interview
witnesses/suspects, verify alibis).

\section{Behavioral Feature Definitions}
\label{app:features}

All 16 features are organized into six categories:

\paragraph{Speed \& Efficiency (2).} \texttt{avg\_turns\_per\_stage}: mean
turns per completed stage. \texttt{exploration\_efficiency}: unique
locations / total moves.

\paragraph{Information Gathering (4).} \texttt{read\_ratio},
\texttt{talk\_ratio}, \texttt{test\_ratio}: proportion of core actions
devoted to each type. \texttt{info\_depth}: information actions / unique
sources.

\paragraph{Exploration Pattern (3).} \texttt{move\_ratio}: proportion of
core actions that are moves. \texttt{location\_coverage}: unique locations
/ 6. \texttt{revisit\_rate}: revisits / total visits.

\paragraph{Social Behavior (2).} \texttt{social\_breadth}: unique
characters / available characters. \texttt{social\_dependency}: talk
actions / information actions.

\paragraph{Decision Patterns (3).} \texttt{action\_diversity}: unique
action types / available types. \texttt{decision\_consistency}: $1 -
\text{CV}(\text{turns per stage})$. \texttt{hesitation\_index}: location
revisits / moves.

\paragraph{Thoroughness (2).} \texttt{resource\_coverage}: unique resources
/ total available. \texttt{experimentation\_rate}: test actions /
information actions.

\section{Action-Word List}
\label{app:aw_list}

The action-word (AW) baseline (\S\ref{sec:baseline}) counts occurrences of
the following terms in each persona description: \textit{read},
\textit{reads}, \textit{conversation}, \textit{discussion}, \textit{test},
\textit{testing}, \textit{tests}, \textit{verify}, \textit{verifies},
\textit{examine}, \textit{check}, \textit{cross-check}, \textit{hands-on},
\textit{practical}, \textit{doing}, \textit{systematic}, \textit{thorough},
\textit{comprehensive}.

\section{Abstract Persona Rewrite}
\label{app:abstract_rewrite}

Section~\ref{sec:causal_anchors} compares the original
\textit{social\_collaborative} persona with an abstract rewrite. Below
we provide the full prompt components for both versions.

\paragraph{Original -- Personality Description.}
``Strongly prefers learning through conversation and discussion. Naturally
seeks others' perspectives, starts by getting oriented through
conversation before consulting other resources. Understands technical
details eventually require written sources.''

\paragraph{Original -- Stage Reminder.}
``Start with conversations to get oriented, then use other resources for
specific details.''

\paragraph{Abstract -- Personality Description.}
``Values collaborative epistemic exchange and dialogic sense-making.
Gravitates toward intersubjective knowledge construction, preferring to
establish shared understanding before engaging with static information
sources. Recognizes that formalized knowledge eventually requires
engagement with codified materials.''

\paragraph{Abstract -- Stage Reminder.}
``Initiate through collaborative sense-making to establish orientation,
then engage with codified sources for specific elaboration.''

\noindent The abstract version preserves the same cognitive disposition
(collaborative, shared-understanding-first, written-sources-second) while
removing all concrete action words (\textit{conversation},
\textit{discussion}, \textit{talk}) that could directly map to the
\texttt{talk} action type. The task instructions component (game rules,
available actions, structural constraints) is identical across both
versions.

\section{Per-Model \bci{} Tables}
\label{app:bci}

Full pairwise correlations for each model (format: \ci{}-\sd{},
\ci{}-\mm{}, \sd{}-\mm{}, \bci{}):

\noindent\textbf{LLaMA-3.1-70B.} 

\vspace{0.3cm}
\textit{social\_collab.}: +.82, +.76, +.81, \textbf{+.80}$^\dagger$. \\
\textit{confident\_dec.}: +.69, +.81, +.69, +.73$^\dagger$. \\
\textit{hands\_on\_prac.}: +.64, +.82, +.48, +.65$^\dagger$. \\
\textit{method.\_thor.}: +.56, +.61, +.54, +.57$^\dagger$. \\
\textit{hypoth.\_driv.}: +.57, +.26, +.48, +.44$^\dagger$. \\

\noindent\textbf{GPT-4o.}

\vspace{0.3cm}
\textit{effortful\_learn.}: +.32, $-$.09, +.61, +.28. \\
\textit{hands\_on\_prac.}: +.29, +.34, +.19, +.27. \\
\textit{social\_collab.}: $-$.23, +.13, +.76, +.22. \\
\textit{quick\_intuit.}: $-$.00, +.37, +.23, +.20. \\
\textit{hypoth.\_driv.}: $-$.02, $-$.31, $-$.21, $-$.18. \\

\noindent\textbf{GPT-4o-OSS-120B.}

\vspace{0.3cm}
\textit{quick\_intuit.}: +.62, +.35, +.61, +.52$^\dagger$. \\
\textit{verif.\_seek.}: +.19, +.26, +.44, +.30. \\
\textit{social\_collab.}: +.58, $-$.03, +.32, +.29. \\
\textit{big\_pict.\_conc.}: +.02, $-$.02, $-$.59, $-$.20. \\
\textit{hands\_on\_prac.}: $-$.20, +.14, $-$.57, $-$.21. \\

\section{Additional Results}
\label{app:additional}

\begin{table}[h!]
\centering\small
\begin{tabular}{llccc}
\toprule
\textbf{Model} & \textbf{Target} & \textbf{All 10} & \textbf{Top 5} & $\boldsymbol{\Delta}$ \\
\midrule
\multirow{3}{*}{LLaMA}
& \ci{}  & 11.4\% & 21.4\% & +10.0 \\
& \sd{}  & 10.2\% & 19.5\% & +9.3  \\
& \mm{}  & 13.6\% & 23.8\% & +10.2 \\
\midrule
\multirow{3}{*}{GPT-4o}
& \ci{}  & 11.0\% & 21.4\% & +10.5 \\
& \sd{}  & 10.5\% & 15.7\% & +5.2  \\
& \mm{}  & 9.8\%  & 20.5\% & +10.7 \\
\midrule
\multirow{3}{*}{GPT-OSS}
& \ci{}  & 9.5\%  & 19.0\% & +9.5  \\
& \sd{}  & 9.3\%  & 34.8\% & +25.5 \\
& \mm{}  & 11.4\% & 28.1\% & +16.7 \\
\midrule
\textbf{Average} & & 10.7\% & \textbf{22.7\%} & \textbf{+12.0} \\
\bottomrule
\end{tabular}
\caption{Persona selection using source-narrative \bci{} only (no target
data). The selected subset improves transfer identifiability in all 9
conditions.}
\label{tab:selection}
\end{table}

\begin{table}[h!]
\centering
\begin{minipage}[t]{0.46\textwidth}
\centering\small
\begin{tabular}{lcc}
\toprule
\textbf{Narrative} & \textbf{\texttt{talk}} & \textbf{Domain} \\
\midrule
\ci{} & .286 & Medical \\
\ck{} & .213 & Cooking \\
\mm{} & .183 & Crime \\
\sd{} & .062 & IT \\
\midrule
Mean & .177 & --- \\
\bottomrule
\end{tabular}
\captionof{table}{\texttt{talk} ratio by narrative (GPT-4o). \ck{}
exceeds the 3-game mean ($d{=}0.48$, $p{<}.001$).}
\label{tab:cooking_social}
\end{minipage}\hfill
\begin{minipage}[t]{0.50\textwidth}
\centering\small
\begin{tabular}{lccc}
\toprule
\textbf{Persona} & \textbf{\bci{}-3} & \textbf{\bci{}-4} & $\boldsymbol{\Delta}$ \\
\midrule
social\_col.     & +.22 & +.25 & +.03 \\
effortful\_l.    & +.28 & +.24 & $-$.04 \\
hands\_on\_p.    & +.27 & +.23 & $-$.04 \\
quick\_int.      & +.20 & +.22 & +.02 \\
method.\_th.     & +.12 & +.14 & +.02 \\
\midrule
verif.\_seek.    & +.03 & +.07 & +.04 \\
cover.\_foc.     & $-$.00 & +.02 & +.02 \\
confid.\_dec.    & $-$.06 & $-$.03 & +.03 \\
big\_pic.\_c.    & $-$.04 & $-$.08 & $-$.04 \\
hypoth.\_dr.     & $-$.18 & $-$.14 & +.04 \\
\bottomrule
\end{tabular}
\captionof{table}{\bci{} over 3 vs.\ 4 narratives. Rankings stable
($\rho{=}0.952$); top-5 identical.}
\label{tab:bci_4game}
\end{minipage}
\end{table}

\begin{table}[h!]
\centering\small
\begin{tabular}{llccc}
\toprule
\textbf{Feature} & \textbf{Model} & \textbf{\ci{}} & \textbf{\sd{}} & \textbf{\mm{}} \\
\midrule
\multirow{3}{*}{\texttt{read\_ratio}}
& LLaMA   & .21 & \textbf{.52} & .36 \\
& GPT-4o  & .24 & \textbf{.58} & .39 \\
& GPT-OSS & .27 & \textbf{.63} & .46 \\
\midrule
\multirow{3}{*}{\texttt{talk\_ratio}}
& LLaMA   & \textbf{.39} & .08 & .21 \\
& GPT-4o  & \textbf{.29} & .06 & .18 \\
& GPT-OSS & \textbf{.30} & .06 & .14 \\
\midrule
\multirow{3}{*}{\texttt{test\_ratio}}
& LLaMA   & .03 & .06 & \textbf{.15} \\
& GPT-4o  & .06 & .07 & \textbf{.19} \\
& GPT-OSS & .03 & .04 & \textbf{.14} \\
\bottomrule
\end{tabular}
\caption{Mean action ratios by task narrative and model. Bold marks the
highest value per row. All three models exhibit the same pattern: \sd{}
elicits the most reading, \ci{} the most social interaction
(\texttt{talk}), and \mm{} the most testing.}
\label{tab:game_profiles}
\end{table}

\begin{table}[h!]
\centering\small
\begin{tabular}{llrc}
\toprule
\textbf{Narrative bias} & \textbf{Model} & \textbf{$r$} & \textbf{Opt.?} \\
\midrule
\multirow{3}{*}{\sd{} $\to$ \texttt{read}}
& LLaMA & $+.35$*** & \cmark \\
& GPT-4o & $+.32$*** & \cmark \\
& GPT-OSS & $+.34$*** & \cmark \\
\midrule
\multirow{3}{*}{\ci{} $\to$ \texttt{talk}}
& LLaMA & $+.18$* & $\sim$ \\
& GPT-4o & $-.29$*** & \xmark \\
& GPT-OSS & $-.13$ & \xmark \\
\midrule
\multirow{3}{*}{\mm{} $\to$ \texttt{test}}
& LLaMA & $-.14$* & \xmark \\
& GPT-4o & $+.04$ & \xmark \\
& GPT-OSS & $-.25$*** & \xmark \\
\bottomrule
\end{tabular}
\caption{Correlation between narrative-biased action and task success.
Only \texttt{read} in \sd{} is beneficial; the biases in \ci{} and
\mm{} hurt or do not help performance.}
\label{tab:bias_vs_optimal}
\end{table}

\begin{table}[h!]
\centering\small
\begin{tabular}{lccc}
\toprule
\textbf{Model} & \textbf{All 10} & \textbf{Top 5} & $\boldsymbol{\Delta}$ \\
\midrule
LLaMA-70B      & 11.4\% & 21.4\% & +10.1 \\
GPT-4o         & 10.0\% & 19.4\% & +9.4  \\
GPT-OSS-120B   & 9.8\%  & 25.1\% & +15.3 \\
\midrule
\textbf{Average} & 10.4\% & \textbf{22.0\%} & +11.6 \\
\bottomrule
\end{tabular}
\caption{Leave-one-narrative-out validation. \bci{} computed from two task
narratives; transfer tested on the held-out third.}
\label{tab:logo}
\end{table}

\begin{table}[h!]
\centering\small
\begin{tabular}{llccc}
\toprule
\textbf{Feature} & \textbf{Model} & \textbf{\ci{}} & \textbf{\sd{}} & \textbf{\mm{}} \\
\midrule
\multirow{3}{*}{\texttt{read}}
& LLaMA   & +.30\rlap{***} & +.35\rlap{***} & +.21\rlap{**} \\
& GPT-4o  & +.44\rlap{***} & +.32\rlap{***} & $-$.13 \\
& GPT-OSS & +.24\rlap{***} & +.34\rlap{***} & +.35\rlap{***} \\
\midrule
\multirow{3}{*}{\texttt{talk}}
& LLaMA   & +.18\rlap{*}   & $-$.18\rlap{**} & +.44\rlap{***} \\
& GPT-4o  & $-$.29\rlap{***} & $-$.01 & +.03 \\
& GPT-OSS & $-$.13          & $-$.02 & $-$.15\rlap{*} \\
\midrule
\multirow{3}{*}{\texttt{move}}
& LLaMA   & $-$.43\rlap{***} & $-$.34\rlap{***} & $-$.35\rlap{***} \\
& GPT-4o  & +.05             & $-$.35\rlap{***} & +.08 \\
& GPT-OSS & $-$.06           & $-$.56\rlap{***} & +.05 \\
\bottomrule
\end{tabular}
\caption{Feature--success correlations by task narrative. *\,$p<.05$,
**\,$p<.01$, ***\,$p<.001$.}
\label{tab:feat_success}
\end{table}

\begin{table}[h!]
\centering\small
\begin{tabular}{lccc}
\toprule
& \textbf{LLaMA} & \textbf{GPT-4o} & \textbf{GPT-OSS} \\
\midrule
\ci{}$\to$\sd{}  & 20.0\% & 12.4\% & 7.6\%  \\
\ci{}$\to$\mm{}  & 20.0\% & 20.0\% & 17.1\% \\
\sd{}$\to$\ci{}  & 20.0\% & 21.0\% & 19.0\% \\
\sd{}$\to$\mm{}  & 28.6\% & 21.0\% & 25.7\% \\
\mm{}$\to$\ci{}  & 21.9\% & 23.8\% & 19.0\% \\
\mm{}$\to$\sd{}  & 19.0\% & 15.2\% & 46.7\% \\
\midrule
\textbf{Mean}    & 21.6\% & 18.9\% & 22.5\% \\
\bottomrule
\end{tabular}
\caption{Full cross-narrative transfer accuracy for top-5 \bci{} personas.}
\label{tab:full_transfer}
\end{table}

\begin{table}[h!]
\centering\small
\begin{tabular}{lcccc}
\toprule
\textbf{Model} & $\boldsymbol{\rho_{\bci{}}}$ & \textbf{$p$} & $\boldsymbol{\rho_{\text{AW}}}$ & \textbf{$p$} \\
\midrule
LLaMA-70B    & .76 & .011  & .55 & .10  \\
GPT-4o       & .68 & .031  & .38 & .28  \\
GPT-OSS-120B & .72 & .019  & .41 & .24  \\
\bottomrule
\end{tabular}
\caption{Correlation of \bci{} and AW count with per-persona transfer
accuracy. \bci{} is significant in all models; AW is not.}
\label{tab:aw_corr}
\end{table}

\begin{table}[h!]
\centering\small
\begin{tabular}{lcccc}
\toprule
\textbf{Model} & \textbf{All 10} & \textbf{AW Top-5} & \textbf{\bci{} Top-5} & $\boldsymbol{\Delta}$ \\
\midrule
LLaMA-70B    & 11.3\% & 17.8\% & \textbf{21.6\%} & +3.8 \\
GPT-4o       & 10.0\% & 14.2\% & \textbf{18.9\%} & +4.7 \\
GPT-OSS-120B & 9.8\%  & 16.5\% & \textbf{22.5\%} & +6.0 \\
\midrule
\textbf{Avg.} & 10.4\% & 16.2\% & \textbf{21.0\%} & +4.8 \\
\bottomrule
\end{tabular}
\caption{Transfer accuracy by selection method. \bci{} top-5 outperforms
AW top-5 in all models. Random-5 mean: 13.1\% [10.8\%, 15.6\%].}
\label{tab:aw_selection}
\end{table}

\section{Confusion Matrices}
\label{app:confusion}

Tables~\ref{tab:cm_within}--\ref{tab:cm_transfer} show per-persona
classification accuracy (diagonal) from the within-narrative and
cross-narrative settings described in \S\ref{sec:narrative_dom}. We report
the diagonal (correct classification rate per persona) for compactness;
full $10 \times 10$ matrices are available in our code release.

\begin{table}[h!]
\centering\small
\begin{tabular}{lccc}
\toprule
\textbf{Persona} & \textbf{\ci{}} & \textbf{\sd{}} & \textbf{\mm{}} \\
\midrule
social\_collab.     & 42.9\%          & \textbf{90.5\%} & 54.0\% \\
hands\_on\_prac.    & 36.5\%          & \textbf{63.5\%} & 39.7\% \\
coverage\_focused   & 36.5\%          & 44.4\%          & 14.3\% \\
verification\_seek. & 31.7\%          & 38.1\%          & 25.4\% \\
confident\_dec.     & 33.3\%          & 33.3\%          & 31.7\% \\
methodical\_thor.   & 20.6\%          & 20.6\%          & 28.6\% \\
big\_picture\_conc. & 27.0\%          & 23.8\%          & 22.2\% \\
quick\_intuitive    & 22.2\%          & 25.4\%          & 15.9\% \\
hypothesis\_driv.   & 14.3\%          & 27.0\%          & 19.0\% \\
effortful\_learn.   & 19.0\%          & 27.0\%          & \textbf{0.0\%} \\
\midrule
\textbf{Mean}       & 28.4\%          & 39.4\%          & 25.1\% \\
\bottomrule
\end{tabular}
\caption{Within-narrative per-persona classification accuracy (diagonal of
confusion matrix). Personas with concrete behavioral anchors
(\textit{social\_collaborative}, \textit{hands\_on\_practical}) are most
identifiable. \textit{Effortful\_learner} is unidentifiable in \mm{}.}
\label{tab:cm_within}
\end{table}

\begin{table}[h!]
\centering\small
\begin{tabular}{lcc}
\toprule
\textbf{Direction} & \textbf{Accuracy} & \textbf{Dominant Prediction} \\
\midrule
\ci{} $\to$ \sd{}  & 6.7\%  & social\_collab. (76.3\%) \\
\ci{} $\to$ \mm{}  & 8.6\%  & social\_collab. (83.0\%) \\
\sd{} $\to$ \ci{}  & 10.0\% & social\_collab. (96.7\%) \\
\sd{} $\to$ \mm{}  & 9.8\%  & social\_collab. (71.4\%) \\
\mm{} $\to$ \ci{}  & 9.5\%  & hands\_on\_prac. (35.7\%) \\
\mm{} $\to$ \sd{}  & 10.0\% & effortful\_learn. (29.4\%) \\
\bottomrule
\end{tabular}
\caption{Cross-narrative transfer: overall accuracy and the single
most-predicted class. In 4 of 6 directions, the classifier collapses to
predicting \textit{social\_collaborative} for nearly all inputs, reflecting
the dominant behavioral bias of the target task narrative.}
\label{tab:cm_transfer}
\end{table}

\section{Cooking Kitchen Game Details}
\label{app:cooking}

The Cooking Kitchen (\ck{}) game (\S\ref{sec:cooking}) preserves all
structural invariants of the original three games. Below we summarize its
domain-specific configuration.

\paragraph{Setting.} A kitchen inspector investigates a dish preparation
problem across a professional kitchen facility.

\paragraph{Stages.} (1) Cuisine category $\to$ (2) Specific dish $\to$
(3) Problem source $\to$ (4) Correction.

\paragraph{Locations (6).} Main Kitchen, Prep Station, Cold Storage,
Pantry, Dining Floor, Office.

\paragraph{Action Mapping.}
\texttt{read}: recipe cards, inspection logs, ingredient lists, supplier
documents.
\texttt{talk}: head chef, sous chef, line cooks, servers, kitchen manager.
\texttt{test}: taste dishes, check temperatures, inspect ingredients,
verify storage conditions.
\texttt{move}: navigate between the 6 locations.

\paragraph{Scenarios (7).} Each scenario involves a different cuisine
category (e.g., Italian, Japanese, Mexican) with a unique dish, problem
source, and correction. Scenario content is generated dynamically from
a fixed configuration to ensure structural equivalence with the original
games.

\paragraph{Role.} The agent plays a kitchen inspector. The persona
descriptions are identical to those used in the original three games
(Appendix~\ref{app:personas}), with only the role label and
domain-specific resource names changed.

\end{document}